# Machine Learning Architectures for the Estimation of Predicted Occupancy Grids in Road Traffic

Parthasarathy Nadarajan and Michael Botsch
Technische Hochschule Ingolstadt, Ingolstadt, Germany
Email: {parthasarathy.nadarajan, michael.botsch}@thi.de

Sebastian Sardina
RMIT, Melbourne, Australia
Email: sebastian.sardina@rmit.edu.au

*Abstract*—This paper introduces a novel machine learning architecture for an efficient estimation of the probabilistic space-time representation of complex traffic scenarios. A detailed representation of the future traffic scenario is of significant importance for autonomous driving and for all active safety systems. In order to predict the future space-time representation of the traffic scenario, first the type of traffic scenario is identified and then the machine learning algorithm maps the current state of the scenario to possible future states. The input to the machine learning algorithms is the current state representation of a traffic scenario, termed as the *Augmented Occupancy Grid* (AOG). The output is the probabilistic space-time representation which includes uncertainties regarding the behaviour of the traffic participants and is termed as the *Predicted Occupancy Grid* (POG). The novel architecture consists of two *Stacked Denoising Autoencoders* (SDAs) and a set of Random Forests. It is then compared with the other two existing architectures that comprise of SDAs and DeconvNet. The architectures are validated with the help of simulations and the comparisons are made both in terms of accuracy and computational time. Also, a brief overview on the applications of POGs in the field of active safety is presented.

## I. Introduction

Active vehicle safety systems such as *Autonomous Emergency Braking* (AEB) are intended to support the driver in situations with potential risk of accidents and hence it is necessary to continuously observe and analyse the traffic situations [1]. Modern day exteroceptive sensors such as camera, laser-scanner, radar, etc. are capable of providing detailed information about the driving environment, but significant progress is required in interpreting the collected information in all traffic scenarios correctly. The understanding of the traffic environment is necessary for the prediction of the behaviour of the traffic participants in the respective traffic situation.

Currently, extensive research is being carried out in anticipating the behaviour of the traffic participants for a specific type of traffic situation, e. g., intersections [2]. In order to predict the behaviour, a number of statistical learning approaches such as Dynamic Bayesian Networks [3], Hidden Markov Model [4] and *Random Forests* (RFs) [5] are adopted. In [5], a probabilistic space-time representation of the future traffic scenario termed as the *Predicted-Occupancy Grid* (POG) is formulated. However, it is important to note that these approaches are applicable only for a specific configuration of traffic scenario. Hence, it is necessary to have a generic approach that is capable of extracting relevant information about the traffic environment and to learn and use the acquired information for estimating the behaviour of the traffic participants when a similar scenario is encountered [6]. In [7], a hierarchical situation classifier is designed to classify different kinds of traffic scenarios encountered during driving. Each leaf of the classifier corresponds to a set of similar traffic scenarios and a machine learning algorithm is trained for that particular scenario to predict the behaviour of the traffic participants.

The representation of the traffic environment plays a crucial role in a reliable estimation of the future behaviour of the traffic participants. In [5], an efficient representation of the current state of the traffic scenario termed as the *Augmented Occupancy Grid* (AOG) is introduced. The cells of an occupancy grid are augmented with information about the road infrastructure and the traffic participants to generate an AOG, which leads to a high-dimensional input vector. The performance of a machine learning approach is deteriorated with an increase in the dimensionality of the input data [8]. Hence, it is meaningful to extract low-dimensional features from the high-dimensional input and to use these low-dimensional features as input to the machine learning algorithm. Though a number of dimensionality reduction techniques are available, deep learning approaches such as Autoencoders [9] and Restricted Boltzmann machines [10], are capable of outperforming the standard approaches. In [7], a *Stacked Denoising Autoencoder* (SDA) is used to extract low-dimensional features from AOGs in an unsupervised manner and the extracted features are then used by the RF algorithm to predict the POGs. The estimation of the POGs using the RF algorithm still has a high level of computational complexity. Hence, in this paper, the SDA is also used to perform dimensionality reduction on the POG and the RF algorithm is only used for the mapping of the low-dimensional features of the AOG to the low-dimensional features of the POG. In [11], a deconvolution network is learned to perform semantic pixel-wise segmentation. Since the estimation of the probabilities of POGs can be considered as a pixel-wise classification task, a deconvolution network is trained to predict the POGs. Since an efficient estimation of the POGs is a crucial aspect for its application in vehicle safety, the paper compares the three different machine learning architectures for the estimation of the POGs in terms of the accuracy and execution time.

The paper is organized as follows. Section II introduces the POGs and their application in vehicle safety. The new proposed machine learning architecture and the other two existing architectures for the estimation of the POGs is presented in Section III. In Section IV, the proposed methodologies are evaluated with the help of simulations and video data from the real world.

Throughout this work, vectors and matrices are denoted by lower and upper case bold letters, respectively. A lower case represents a column vector.

## II. PREDICTED OCCUPANCY GRIDS IN VEHICLE SAFETY

One of the challenges faced by the active safety systems is to anticipate the behaviour of the traffic participants in a particular traffic scenario. The task of predicting the behaviour of the traffic participants for different kinds of scenarios in the real world is highly complex. Therefore, the best-suited approach to address such a complex problem is to "Divide and Conquer". A hierarchical situation classifier is designed in [7] to classify the different traffic situations based on the road geometry and to identify the safety-relevant traffic participants in the corresponding scenario. The participants that encounter the EGO vehicle (the vehicle in which the safety algorithm operates) are referred to as the safety-relevant participants. The classification of the road geometry is based on an *Image Distortion Model* (IDM) and the k-*Nearest Neighbour* (k-NN) classifier [12] and the determination of safety-relevant participants is based on a rule-based classifier. Each node of the classifier represent a similar set of traffic scenarios, e. g., straight road, left curve, right curve, cross junction etc. and a machine learning algorithm is trained for each node to predict the behaviour of the safety-relevant traffic participants. An overview of the hierarchical situation classifier can be seen in the Figure 1.

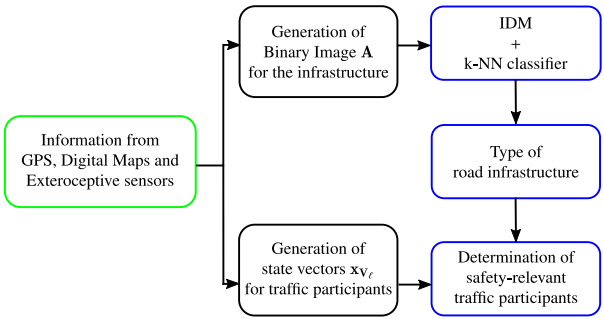

Figure 1. Hierarchical Situation Classifier.

A grid based efficient probabilistic space-time representation of the future traffic scenarios termed as the *Predicted Occupancy Grid* (POG) is introduced in [5]. The main advantage of the POG is that it includes a detailed modelling of the uncertainties regarding the motion behaviour of the traffic participants by considering their multiple motion hypotheses. A POG $\mathcal{G}_{t_\text{pred}}$ is computed for each prediction time instance $t_\text{pred}$. Hence, a time horizon divided into $\kappa$ intervals will result in $\kappa$ POGs. A POG has rows $I$ and $J$ columns and the $(i,j)$-th cell of the POG at prediction time instance $t_\text{pred}$ is $g^{i,j}_{t_\text{pred}}$ and its probability of occupancy is $\text{p}^{i,j}_{t_\text{pred}}$. Since the POG includes multiple hypotheses of the traffic participants, the $\text{p}^{i,j}_{t_\text{pred}}$ depends on how likely the individual trajectories of the traffic participants are. It should be noted that $g^{i,j}_{t_\text{pred}}$ can be simultaneously occupied by multiple trajectory hypotheses of the traffic participants. However, the maximum value of $\text{p}^{i,j}_{t_\text{pred}}$ is limited to 1 as:

$$\text{p}^{ij}_{t_\text{pred}} = \min\left(1, \sum_{\ell=1}^{L} \left((\mathbf{z}^{ij}_{V_\ell, t_\text{pred}})^\text{T} \mathbf{p}(h_{V_\ell, t_\text{pred}})\right)\right), \quad (1)$$

where $\mathbf{z}^{ij}_{V_\ell, t_\text{pred}}$ corresponds to a binary vector of size $S$ and $\ell = 1, \dots, L$ denotes the number of traffic participants. $S$ is the number of trajectory hypotheses per traffic participant. Depending on the occupancy of $g^{i,j}_{t_\text{pred}}$ given by the $S$ multiple hypotheses trajectories of the traffic participant $V_\ell$, $\mathbf{z}^{ij}_{V_\ell, t_\text{pred}}$ takes value 0 or 1. The probabilities of the $S$ multiple hypotheses at time instance $t_\text{pred}$ is given by the vector $\mathbf{p}(h_{V_\ell, t_\text{pred}})$. Though model based approaches can be used for the estimation of POGs, they require high computational effort and are not real-time capable [5]. The estimation of POGs in real-time is of crucial importance for its application in vehicle safety and thus, a machine learning approach is adopted.

The applications of the POGs in the field of vehicle safety is explained in Section II-A. Section II-B introduces the AOG, i.e., the current state representation of the traffic scenario.

### A. Vehicle Safety Applications for Collision Mitigation

The detailed modelling of the uncertainties regarding the motion behaviour of the traffic participants in real-time with the machine learning approach helps in improving important components of vehicle safety like criticality estimation and trajectory planning. In critical situations, it is important to plan a trajectory for the EGO vehicle which has a very low risk of collision with the surrounding traffic participants. Since the POGs consider multiple hypotheses of the traffic participants, current trajectory planning approaches such as *Rapidly-exploring Random Tree* (RRT) algorithm [13] can be combined with the POGs for finding safe trajectories with a very low risk of collision and at the same time maintain a fast computational speed.

With increasing complexity of the active safety systems, conventional testing procedures like Software-in-the-loop, Hardware-in-the-loop, test drives etc., are insufficient. It is important to note that it is practically infeasible to test systems through an infinite range of traffic scenarios. Hence, the choice of test scenarios for validating a system is one important aspect of the testing process. POGs offer the possibility to reduce the number of test scenarios by clustering scenarios based on the hypotheses about the future development of a traffic scenario. Such a reduction in the number of test scenarios contribute towards a significant reduction of the resources and will be the context of future work.

### B. Augmented Occupancy Grid (AOG)

As stated earlier, an efficient representation of the current state of the traffic scenario is crucial for the estimation of the POGs using the machine learning approach. Since the evolution of a traffic scenario depends on the intention of the traffic participants and the interaction between them, the current state representation should contain information about the road infrastructure and the traffic participants. A novel method to represent the current state of the traffic scenario with all the necessary information is introduced in [5] and is termed as the *Augmented Occupancy Grid* (AOG) $\mathcal{OG}_0$. It is important to note that a specific traffic scenario has one AOG $\mathcal{OG}_0$, with the subscript 0 denoting the current time instance $t_0$. An occupancy grid is generated from the traffic scenario under observation at $t_0$. The occupancy grid has $I$ columns and $J$ rows, with $\ell_{\text{cell}}$ and $w_{\text{cell}}$ being the length and width of the cells, respectively. The cells of the occupancy grid are then augmented with information about the road infrastructure and traffic participants to generate the AOG. If a traffic participant $V_\ell$ with the velocity $v_\ell$, orientation $\psi_\ell$, longitudinal and lateral acceleration $a_{x,\ell}$ and $a_{y,\ell}$ respectively occupies a cell of the occupancy grid, the corresponding attributes of the cells in the $\mathcal{OG}_0$ are $[1, v_\ell, \psi_\ell, a_{x,\ell}, a_{y,\ell}]^{\text{T}}$. Similarly, the attributes of the cells in the $\mathcal{OG}_0$ for a stationary object and road infrastructures like lane markings are $[1, 0, 0, 0, 0]^{\text{T}}$.

### III. MACHINE LEARNING ARCHITECTURES FOR THE ESTIMATION OF POG

The estimation of the POGs has to be efficient both in terms of accuracy and computation time for its application in vehicle safety. With the machine learning approach offering a solution to solve this complex task with high accuracy, it is important to choose a suitable architecture of the machine learning for the computation of the POGs.

This section deals with the theoretical background of the various machine learning algorithms for the computation of POGs followed by the description of the three different machine learning architectures.

### A. Stacked Denoising Autoencoder (SDA)

In general, it is a challenge to deal with high dimensional input space when performing machine learning tasks due to the curse of dimensionality. In this work, a successful unsupervised technique for dimensionality reduction, the SDA is used.

A neural network with three layers viz., the input, hidden and reconstruction layer, that is trained to learn its input, is an autoencoder. It consists of an encoder and decoder function that is responsible for mapping the input to the hidden layer and the hidden to the input layer, respectively. With minimal reconstruction error, it can be stated that the hidden layer is a low-dimensional representation of the input vector. A corrupting operation is performed on the input as a regularization measure to prevent the autoencoder from learning the identity function and such autoencoders are termed as denoising autoencoders. In order to learn better representations of the input vector, multiple denoising autoencoders are stacked one above the other to form the SDA and a single layer of the SDA can be seen in the Figure 2.

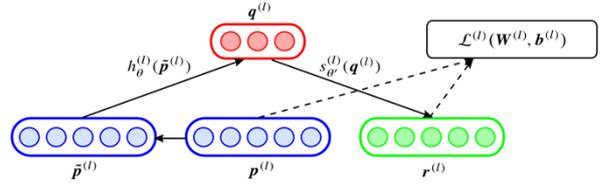

Figure 2. Single Layer of Stacked Denoising Autoencoder (SDA).

Let $l = 1, \dots, nl$ correspond to the layer of the SDA, with $\boldsymbol{p}^{(l)}$, $\boldsymbol{q}^{(l)}$ and $\boldsymbol{r}^{(l)}$ as the visible, hidden and reconstructed vector of the $l$-th layer, respectively. Corrupting operations such as Gaussian noise, masking noise or salt and pepper noise are added to $\boldsymbol{p}^{(l)}$ to create a partially destroyed version of the input $\widetilde{\boldsymbol{p}}^{(l)}$. The construction of the hidden layer vector $\boldsymbol{q}^{(l)}$ is followed by the corrupting operation using the encoding function $h_\theta^{(l)}(\widetilde{\boldsymbol{p}}^{(l)})$:

$$\boldsymbol{q}^{(l)} = \boldsymbol{h}_\theta^{(l)}(\widetilde{\boldsymbol{p}}^{(l)}) = f(\boldsymbol{W}^{(l)}\widetilde{\boldsymbol{p}}^{(l)} + \boldsymbol{b}^{(l)}), \qquad (2)$$

where $\theta^{(l)} = \{\boldsymbol{W}^{(l)}, \boldsymbol{b}^{(l)}\}$, with $\boldsymbol{W}^{(l)}$ and $\boldsymbol{b}^{(l)}$ being the weight and bias vector of the $l$-th layer, respectively. The function $f(\cdot)$ corresponds to the activation functions such as sigmoid, linear, hyperbolic tangent, etc. The estimation of the reconstructed input vector $\boldsymbol{r}^{(l)}$ from the hidden layer using the decoding function $s_{\theta'}^{(l)}(\boldsymbol{q}^{(l)})$ is given by the following equation:

$$\boldsymbol{r}^{(l)} = \boldsymbol{s}_{\theta'}^{(l)}(\boldsymbol{q}^{(l)}) = f(\boldsymbol{W}^{(l)'}\boldsymbol{q}^{(l)} + \boldsymbol{b}^{(l)'}), \qquad (3)$$

where $\theta^{(l)'} = \{\boldsymbol{W}^{(l)'}, \boldsymbol{b}^{(l)'}\}$. In general, tied weights and bias, i.e., $\boldsymbol{W} = \boldsymbol{W}'$ and $\boldsymbol{b} = \boldsymbol{b}'$ are used for training the autoencoders and a similar approach is adopted in this paper. The loss function $\mathcal{L}^{(l)}$ of the $l$-th layer for the reconstruction of the input is a second-order loss function with regularization parameters and is given by:

$$\mathcal{L}^{(l)}(W^{(l)}, b^{(l)}) = \frac{1}{2G} \sum_{g=1}^{G} \|r_g^{(l)} - p_g^{(l)}\|^2 + \frac{\lambda}{2} \sum_{l=1}^{nl} \sum_{x=1}^{sl} \sum_{y=1}^{sl+1} (W_{xy}^{(l)})^2, \quad (4)$$

where $g = 1, \ldots, G$ is the total number of training data, $\lambda$ is the weight decay parameter, $sl$ represents the number of units on the $l$-th layer and $W_{xy}^{(l)}$ is the weight associated with the connection between unit $x$ in layer $l$ and $y$ in layer $l + 1$.

*B. Random Forest (RF)*

The RF algorithm is used for the mapping of the low-dimensional features of the AOG that result from the SDA to the POG for *Architecture I* and for the mapping of the low dimensional features of the AOG to the low dimensional features of POG as in *Architecture II*. The process is explained in detail in Sections III-D and III-E. The RF algorithm is introduced by Breiman in [14]. The RF algorithm is an ensemble method, where the construction of classification or regression tress are changed in addition to constructing each tree using a different bootstrap sample of the data. In standard trees, each node is split based on the split among all variables. However, in a RF algorithm, each node is split using the best among a sub set of predictors randomly chosen at that node. Such a construction results in a large number of simple classifiers with low bias. By taking a majority vote among the individual classifiers, the variance of the RF classifier is reduced without increasing its bias in comparison to the bias of the individual classifiers in the ensemble [15]. In addition, the main reasons for using the RF algorithm in this paper are its well-known properties such as: good generalization, low number of hyper-parameters to be tuned during training and good performance with high-dimensional data.

*C. Deconvolution Network*

The estimation of probabilities of the POG $\mathcal{G}_{t_{pred}}$ can be considered as a pixel-wise classification task. Though a number of approaches exist for performing pixel-wise classification task, the deconvolution network has recently gained significant popularity due to its ability to work efficiently with image data. The trained network consists of two parts: a convolution and deconvolution network. The convolution network is responsible for the extraction of features and transforms the input vector into a multi-dimensional feature representation. The deconvolution network is a shape generator that generates object segmentation from the extracted features [16]. The output of the final decoder is fed in as input to a trainable soft-max classifier. Each pixel is then classified independently by the soft-max classifier.

Let $M^{(u)}$ be the input for the $u$-th convolution layer with the width $I$, height $J$ and depth $D$ such that $M^{(u)} \in \mathbb{R}^3$. A filter (kernel) $K^{(u)}$ can be defined such that $K^{(u)} \in \mathbb{R}^3$ of size $k_1 \times k_2 \times D$. The output of the $u$-th convolution layer viz. the feature map $FM^{(u)}$ is:

$$FM^{(u)} = f\left(\sum_{m=1}^{k_1} \sum_{n=1}^{k_2} \sum_{d=1}^{D} K_{m,n,d}^{(u)} \cdot M_{i+m,j+n,d}^{(u)} + b^{(u)}\right), \quad (5)$$

where $b^{(u)}$ is the bias for the $u$-th layer and $f(\cdot)$ corresponds to the activation function. Deconvolution, also called as transposed convolution, is done by swapping the forward and backward passes of an input-strided convolution. The deconvolution can also be described as the gradient of a convolution with respect to the input [17]. In the Figure 3, the convolution could be represented by a sparse matrix $S$, where the input matrix is converted to a vector and $K_{k_1,k_2}$ represent the non-zero elements of the sparse matrix $S$. The sparse matrix for the convolution operation shown in the Figure 3 is:

$$\begin{pmatrix} K_{0,0} & K_{0,1} & 0 & K_{1,0} & K_{1,1} & 0 & 0 & 0 & 0 \\ 0 & K_{0,0} & K_{0,1} & 0 & K_{1,0} & K_{1,1} & 0 & 0 & 0 \\ 0 & 0 & 0 & K_{0,0} & K_{0,1} & 0 & K_{1,0} & K_{1,1} & 0 \\ 0 & 0 & 0 & 0 & K_{0,0} & K_{0,1} & 0 & K_{1,0} & K_{1,1} \end{pmatrix}.$$

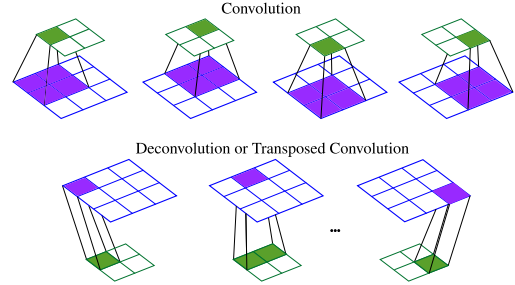

Figure 3. *Top*: Convolving a 2 × 2 kernel over a 3 × 3 input using unit strides. *Bottom*: The transpose of convolving a 2 × 2 kernel over a 3 × 3 using unit strides.

The forward pass takes the input matrix as a 9-dimensional vector and produces a 4-dimensional vector, which upon reshaping result in a 2 × 2 output. The backward pass can be obtained by transposing the sparse matrix, i. e., $S^T$. The operation produces a 9-dimensional output using a 4-dimensional vector as input. Similarly, the forward and backward passes of the deconvolution or transposed convolution, as shown in the Figure 3, are computed by multiplying with $S^T$ and $(S^T)^T = S$, respectively. The output of the $v$-th layer of deconvolution is given by $RM^{(v)}$. The deconvolution is followed by the soft-max layer. Since per-pixel estimation is done, the output of the $(i, j)$-th cell is given by:

$$\hat{y}_{(i,j,c)} = \frac{e^{z_{(i,j,c)}}}{\sum_{k=1}^{C} e^{z_{(i,j,k)}}}, \quad (6)$$

where $z_{(i,j,c)}$ is the log-probability of class $c$ and $C$ is the total number of classes. In order to train the network, cross-entropy loss function is used. With $y_{(i,j,k)}$ being the true class, the loss function is given by:

$$\mathcal{L}_{ce}(i,j) = -\sum_{k=1}^{C} y_{(i,j,k)} \log(\hat{y}_{(i,j,k)}). \quad (7)$$

of RFs to be trained also increases. For example, if the POG is of size $80 \times 80$, then 6400 RFs should be trained and as well stored.

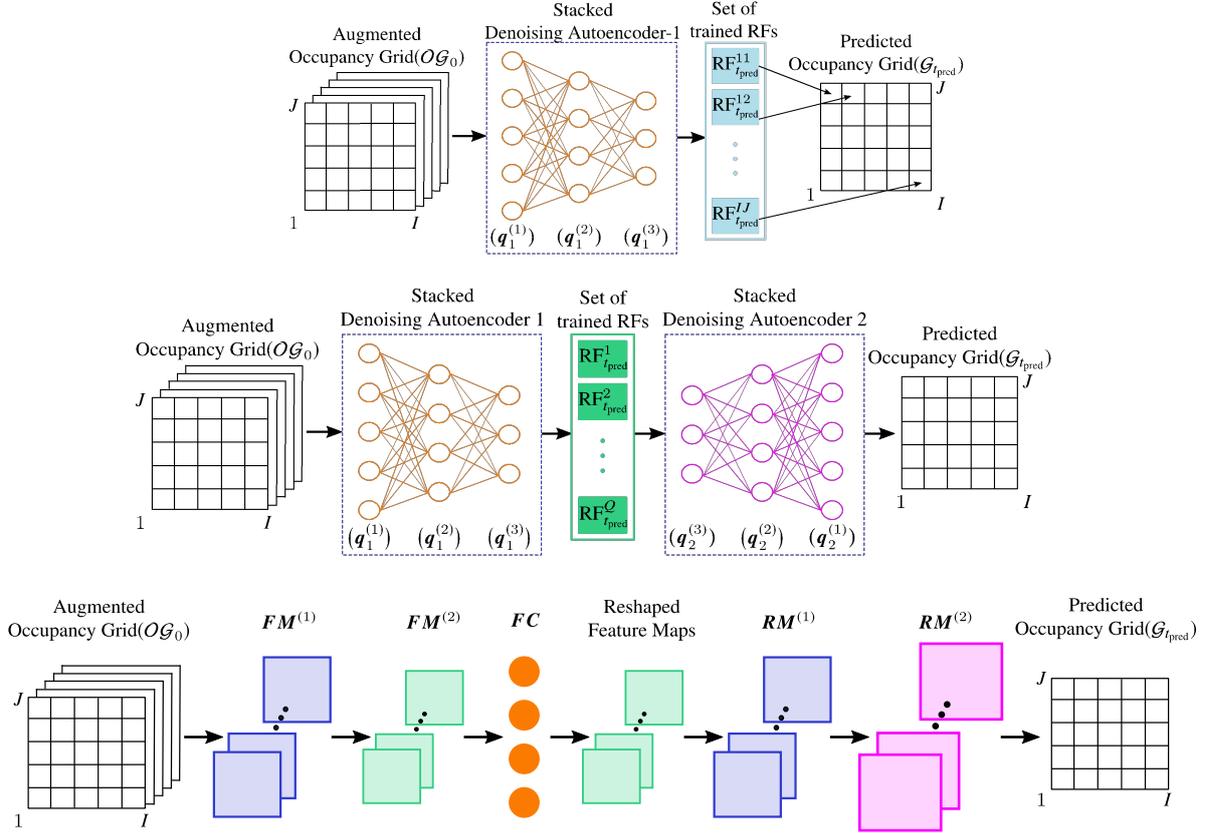

Figure 4. Estimation of POGs using *Top: Architecture I; Middle: Architecture II; Bottom: Architecture III*

### D. Architecture I

An architecture for the estimation of the POGs is presented in [7]. As a first step, a three-layer SDA-1 ($nl = 3$) is used for extracting low-dimensional features $q_1^{(3)}$ from the AOG $\mathcal{OG}_0$. Hence, $q_1^{(1)} = \mathcal{OG}_0$ and $r_1^{(1)}$ corresponds to the reconstructed AOG. The RF-1 algorithm is used in the second step to map the low-dimensional representation $q_1^{(3)}$ to the POG $\mathcal{G}_{t_{\text{pred}}}$. The machine learning algorithms SDA-1 and RF-1 are responsible for the mappings:

$$\begin{aligned} \text{SDA-1:} & \ \mathcal{OG}_0 \mapsto q_1^{(3)} \\ \text{RF-1:} & \ q_1^{(3)} \mapsto \mathcal{G}_{t_{\text{pred}}}, \end{aligned} \quad (8)$$

respectively. The number of hidden neurons in the first, second and third layer of the SDA-1 are 7000, 5000 and 2000, respectively. An overview of the architecture can be seen in the top of Figure 4. Since the cells of the POG are assumed to be independent of each other, one RF is trained per cell of the POG $g_{t_{\text{pred}}}^{i,j}$. Hence, for a prediction time instance $t_{\text{pred}}$, a set of trained RFs $\{\text{RF}_{t_{\text{pred}}}^{11}, \ldots, \text{RF}_{t_{\text{pred}}}^{IJ}\}$ is used to estimate the probabilities $p_{t_{\text{pred}}}^{ij}$ of the POG $\mathcal{G}_{t_{\text{pred}}}$. The main drawback of this approach is that depending on the grid resolution of the POGs, the corresponding number

### E. Architecture II

Though the *Architecture I* accounts for high accuracy, it still requires high computational effort. The number of RFs to be trained for each prediction time instance $t_{\text{pred}}$ is also high. Hence in this work, a new architecture is proposed in which a dimensionality reduction is also performed on the POG using a second SDA. A three layer SDA-2 (with $nl = 3$) is used for extracting low-dimensional features $q_2^{(3)}$ from the POG $\mathcal{G}_{t_{\text{pred}}}$. Hence, $q_2^{(1)} = \mathcal{G}_{t_{\text{pred}}}$ and $r_2^{(1)}$ corresponds to the reconstructed POG. The number of hidden neurons in the first, second and third layer are 5000, 3000 and 2000, respectively. The $q_2^{(3)}$ is a vector of size $Q = 2000$. A set of trained RFs $\{\text{RF}_{t_{\text{pred}}}^{1}, \ldots, \text{RF}_{t_{\text{pred}}}^{Q}\}$ are used to perform the mapping from the low-dimensional representation of the AOG $q_1^{(3)}$ to the low-dimensional representation of the POG $q_2^{(3)}$. An overview of the *Architecture II* can be seen in the middle of Figure 4. The steps involved in the mapping of the AOG $\mathcal{OG}_0$ to the POG $\mathcal{G}_{t_{\text{pred}}}$ are:

$$\begin{aligned} \text{SDA-1:} & \ \mathcal{OG}_0 \mapsto q_1^{(3)} \\ \text{RF-2:} & \ q_1^{(3)} \mapsto q_2^{(3)} \\ \text{SDA-2:} & \ q_2^{(3)} \mapsto \mathcal{G}_{t_{\text{pred}}}. \end{aligned} \quad (9)$$

It is important to note that the encoder part of the SDA-1 is used, whereas in the SDA-2 the decoder part is used. By this architecture, a significant reduction in the number of RFs to be trained for the mapping can be achieved. This in turn helps in the reduction of the computational effort and also in the storage of the trained RFs.

*F. Architecture III*

The DeconvNet used in this paper has an encoder network and a corresponding decoder network followed by a pixel-wise classification layer. The architecture is illustrated in the bottom of Figure 4. The encoder network has 2 convolutional layers. The convolution is done with a stride of 2 to reduce the spatial dimension. The convolution at each layer is followed by the application of the activation function *Rectified Linear Unit* (ReLU). The first and second convolution result separately in $a = 1, \ldots, A$ feature maps $\boldsymbol{FM}_a^{(1)} \in \mathbb{R}^2$ of size $40 \times 40$ and $\boldsymbol{FM}_a^{(2)} \in \mathbb{R}^2$ of size $20 \times 20$, respectively. In this work, the number of filters in each convolution layer is 20, i.e., $A = 20$. It is followed by a fully-connected layer resulting in the output $\boldsymbol{FC}$, which is a 8000 dimension vector. The fully connected layer is then followed by the decoder network. It comprises of 2 deconvolution layers resulting in the outputs $\boldsymbol{RM}_b^{(1)} \in \mathbb{R}^2$ of size $40 \times 40$ and $\boldsymbol{RM}_d^{(2)} \in \mathbb{R}^2$ of size $80 \times 80$, with $b = 1, \ldots, B$ and $d = 1, \ldots, D$, respectively. $B$ and $D$ take up the values 20 and 5, respectively. The final layer is the soft-max layer to estimate the POG $\mathcal{G}_{t_{\text{pred}}}$.

## IV. EVALUATION OF THE METHODOLOGIES

This section describes the evaluation of the machine learning approaches by using appropriate quality metrics. Data from a video dataset is also generated for the validation of the proposed methodologies.

*A. Generation of Data*

In order to validate the machine learning approaches, both the training and validation data is generated with the help of the simulation as described in [5]. The data from the simulation environment is transformed and represented in such a way that the traffic scenario under observation is seen from the perspective of the EGO vehicle. A four-way intersection over a span of $40\text{m} \times 40\text{m}$ with two cars (green) and a bicyclist (blue) along with the EGO vehicle (red). The centre of gravity of the EGO vehicle is located at $(2.5\text{m}, 0\text{m})$. Two different configurations of the traffic participants are considered. In the first configuration, seen in the left side of the Figure 5, the intersection allows traffic in all four directions. The possible manoeuvres of the traffic participants is indicated by the arrow. In the second configuration, the road to the left of EGO vehicle is a no-entry. The corresponding possible manoeuvres of the traffic participants are indicated by the arrows that can be seen in the right side of the Figure 5. The grid resolution of the POG is considered to be 0.5m, thereby resulting in a POG of dimension $80 \times 80$ for a prediction time instance $t_{\text{pred}}$.

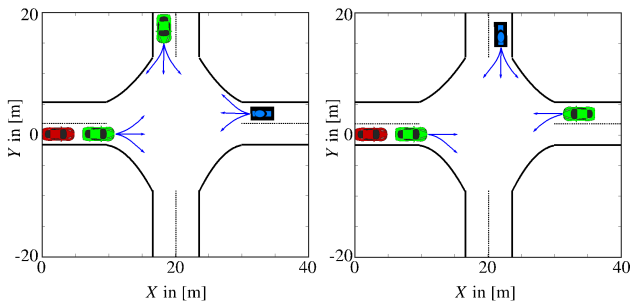

Figure 5. Traffic Scenario under Consideration for the validation.

The dataset comprises of 33280 different traffic scenarios, which are generated by varying the positions, speed, constellations and accelerations of the traffic participants. The position of the traffic participants are varied over a range of 10m. The velocity is varied between 10km/h and 35km/h. The maximum velocity is chosen such that it is realistic for the vehicles to turn at the intersections. The orientation of the traffic participants is changed over a range of 60 degrees. The longitudinal acceleration is varied between $0\text{m/s}^2$ and $4\text{m/s}^2$. A prediction time horizon $t_{\text{pred}}$ of 1.0s is chosen. An example of the POG $\mathcal{G}_{t_{\text{pred}}}$ for the traffic scenario under consideration at $t_{\text{pred}} = 1.0\text{s}$ can be seen in the left hand side of the Figure 6. In order to perform the classification task, the probability of occupancy of the POG cell $\text{p}_{t_{\text{pred}}}^{ij}$ is quantized as:

$$\text{p}_{q,t_{\text{pred}}}^{ij} \in \{0, 0.2, 0.4, 0.6, 0.8, 1.0\}. \qquad (10)$$

The quantization is performed based on the values of the probability of occupancy $\text{p}_{t_{\text{pred}}}^{ij}$. For example, if the $\text{p}_{t_{\text{pred}}}^{ij}$ has a value in the interval $(0.1, 0.2]$ the quantized probability $\text{p}_{q,t_{\text{pred}}}^{ij}$ will be 0.2. The POG after quantization can be seen in the right side of the Figure 6. A total of 24280 traffic scenarios is chosen at random for the training and the remaining 9000 traffic scenarios are used for the validation.

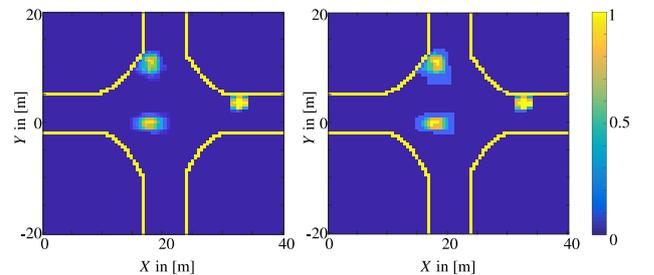

Figure 6. Left: POG for $t_{\text{pred}} = 1.0s$ with color bar denoting the probability of occupancy $\text{p}_{t_{\text{pred}}}^{ij}$. Right: POG for $t_{\text{pred}} = 1.0s$ with color bar denoting the probability of occupancy $\text{p}_{q,t_{\text{pred}}}^{ij}$.

*B. Quality metrics*

Appropriate quality metrics are important for evaluating the performance of the machine learning algorithms. In [7], two separate metrics for evaluating both the performance of the SDA and RF are introduced and the same metrics are adopted in this paper. In order to measure the performance of the SDA, the *Root Mean Squared*

$$\varepsilon_{\text{SDA}} = \|r^{(1)} - p^{(1)}\|. \quad (11)$$

*Error* (RMSE) is used. The error of one data (AOG $\mathcal{OG}_0$ or POG $\mathcal{G}_{t_{\text{pred}}}$) is given by:

Since the estimation of the probability of occupancy in the POG $\mathcal{G}_{t_{\text{pred}}}$ is a crucial task, it is meaningful to use a strict measure which is not rewarding the estimation of free space. Let the non-zero values in the ground truth POG $\mathcal{G}_{t_{\text{pred}}}$ and estimated POG $\hat{\mathcal{G}}_{t_{\text{pred}}}$ be denoted by the set of cells $\mathcal{B}$ and $\mathcal{D}$, respectively. The cardinality $\mathcal{K}$ of the set $(\mathcal{B} \cup \mathcal{D}) \setminus (\mathcal{B} \cap \mathcal{D})$ is given by,

$$\mathcal{K} = |(\mathcal{B} \cup \mathcal{D}) \setminus (\mathcal{B} \cap \mathcal{D})|. \quad (12)$$

Hence, the performance of the three machine learning architectures in predicting the POG $\mathcal{G}_{t_{\text{pred}}}$ for the time instance $t_{\text{pred}}$ is measured as,

$$\epsilon = \sqrt{\frac{1}{\mathcal{K}} \sum_{i=1}^{I} \sum_{j=1}^{J} \left(\hat{p}_{t_{\text{pred}}}^{ij} - p_{t_{\text{pred}}}^{ij}\right)^2}, \quad (13)$$

where $\hat{p}_{t_{\text{pred}}}^{ij}$ and $p_{t_{\text{pred}}}^{ij}$ denote the probabilities stored in the $(i,j)$-th cell of the estimated and ground truth POG, respectively.

### C. Simulation Results

This section presents the results of the simulation performed to evaluate the machine learning architectures.

*1) Dimensionality Reduction:* The performance of the SDA in performing the dimensionality reduction on both the AOG $\mathcal{OG}_0$ and POG $\mathcal{G}_{t_{\text{pred}}}$ is discussed here. The AOG $\mathcal{OG}_0$ of dimension 32000 is reduced to a dimension of 2000 with the help of the SDA-1. Similarly, the SDA-2 is responsible for reducing the dimension of the POG $\mathcal{G}_{t_{\text{pred}}}$ from 6400 to 2000. The hyperparameters of the SDA-1 and SDA-2 are chosen according to [8], in which a detailed analyses of the influence of each parameter on the feature extraction is discussed. For both the SDAs, the hyperparameters are: 1) Number of training epochs: 500; 2) Transfer function: Pure Linear; 3) Corruption operation: Gaussian Noise; 4) Learning Rate: 0.001; 5) Weight decay parameter $\lambda$: 0.005. The SDAs are trained in the Matlab environment using the 24280 generated scenarios as explained in Section IV-A. The performance of the SDAs is estimated based on the error $\varepsilon_{\text{SDA}}$ calculated according to the Equation (11). An overview of the results can be seen in the TABLE I. The first row shows the average absolute value per cell of the AOG $\mathcal{OG}_0$ and the mean value of the RMSE $\bar{\varepsilon}_{\text{SDA-1}}$ estimated over the 9000 test scenarios. The average absolute value per cell of the POG $\mathcal{G}_{t_{\text{pred}}}$ and the RMSE $\bar{\varepsilon}_{\text{SDA-2}}$ over the 9000 test scenarios can be seen in the row two. By comparing the average absolute value per cell and the RMSE, it can be seen that the low-dimensional features extracted using the SDAs are a robust representation of the input. The histogram of the RMSE $\bar{\varepsilon}_{\text{SDA-1}}$ and $\bar{\varepsilon}_{\text{SDA-2}}$ can be seen in the left and right side of the Figure 7, respectively. The standard deviation of the RMSE error indicates that the reconstructed AOG and POG from SDA-1 and SDA-2 respectively are close to the ground truth.

TABLE I. RMSE ERROR OF THE SDA-1 AND SDA-2

| SDA | Average absolute value per cell | RMSE $\bar{\varepsilon}_{\text{SDA}}$ |
|---|---|---|
| SDA-1 | AOG $\mathcal{OG}_0$: 0.1130 | 0.0041 |
| SDA-2 | POG $\mathcal{G}_{t_{\text{pred}}}$: 0.2300 | 0.0072 |

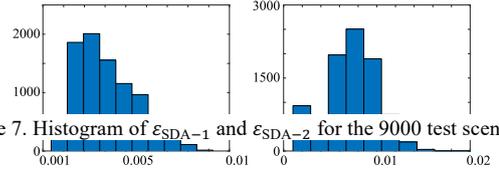

Figure 7. Histogram of $\varepsilon_{\text{SDA}-1}$ and $\varepsilon_{\text{SDA}-2}$ for the 9000 test scenarios

*2) POG Prediction:* The evaluation of the machine learning architectures in the estimation of the POG $\mathcal{G}_{t_{\text{pred}}}$ for the prediction time instance $t_{\text{pred}} = 1.0s$ is presented in this section. The training of the RFs is done in the Matlab environment and the training of the DeconvNet is done in TensorFlow. In order to better analyse the results, three error estimates $\epsilon_{low}$, $\epsilon_{mid}$ and $\epsilon_{high}$ as introduced in [5] for the respective low, mid and high values of probability $p_{t_{\text{pred}}}^{ij}$ are used. The probability ranges for the estimation of $\epsilon_{low}$, $\epsilon_{mid}$ and $\epsilon_{high}$ are $[0, 0.2]$, $(0.2, 0.7]$ and $(0.7, 1.0]$, respectively. The mean error $\bar{\epsilon}$ estimated over the 9000 test scenarios for the three machine learning architectures can be seen in the TABLE II.

TABLE II. ERROR OF THE THREE MACHINE LEARNING ARCHITECTURES BASED ON THE VALIDATION DATASET

| Architecture | $\bar{\epsilon}_{low}$ | $\bar{\epsilon}_{mid}$ | $\bar{\epsilon}_{high}$ |
|---|---|---|---|
| *Architecture I* | 0.0518 | 0.0337 | 0.0277 |
| *Architecture II* | 0.0742 | 0.0739 | 0.0501 |
| *Architecture III* | 0.1501 | 0.1447 | 0.0777 |

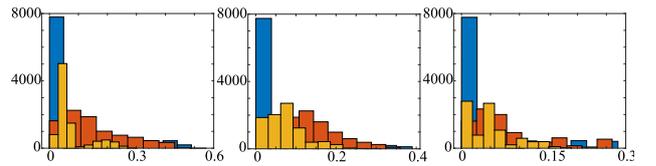

Figure 8. Histogram of $\epsilon_{low}$, $\epsilon_{mid}$ and $\epsilon_{high}$ for 9000 test scenarios for the three machine learning architectures (Blue: *Architecture I*, Yellow: *Architecture II*, Orange: *Architecture III*)

In the Figure 8, the histogram of the three error estimates $\epsilon_{\text{low}}$, $\epsilon_{\text{mid}}$ and $\epsilon_{\text{high}}$ for the three machine learning architectures can be seen. The blue bins correspond to the error estimates for the *Architecture I*. Similarly, the yellow and orange bins correspond to the error estimates for the *Architecture II* and *Architecture III*, respectively. By comparing the results, it can be stated that the highest accuracy in the estimation of the POG $\mathcal{G}_{t_\text{pred}}$ can be achieved with the *Architecture I*, followed by *Architecture II* and then *Architecture III*. In comparison to the *Architecture III*, there is approximately 70% and 50% increase in the accuracy for the *Architecture I* and *Architecture II*, respectively. Since the prediction horizon is $t_\text{pred} = 1.0s$, there is a high probability for the occurrence of mid and high occupancy values. The occurrence of the low occupancy values will have the least probability. This explains the reason for the relatively high error for the low occupancy values in comparison with the mid and high occupancy values. It is important to note that although the *Architecture I* has the highest accuracy in the estimation of the POG $\mathcal{G}_{t_\text{pred}}$, it is computationally very expensive as it involves an SDA and 6400 RFs. The Architecture II requires relatively less computational power as it involves two SDAs and only 2000 RFs. The Architecture III is the least computationally expensive architecture of the three machine learning architectures. Hence, depending on the requirements for the accuracy of the estimation of the POGs along with the available computational power, a corresponding machine learning architecture can be chosen.

*3) Validation with Video Dataset:* Since the validation of the machine learning algorithms is done only with the simulation data, it is important to validate them with some realistic data. Hence, car-bicycle crash compilation videos available in the YouTube are used for generating a new test dataset. The critical scenarios are recreated in the Matlab environment as can be seen in the Figure 9. A total of 80 different scenarios are created with 10 scenarios being 4-way crossing traffic scenarios. Based on the intial states of the traffic participants and the road infrastructure, the AOG $\mathcal{OG}_0$ and the POG $\mathcal{G}_{t_\text{pred}}$ for the 10 crossing scenarios are generated.

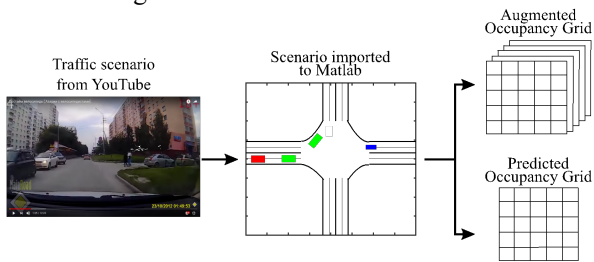

Figure 9. Generation of the traffic scenarios in Matlab using the car-bicycle crash compilation videos from YouTube.

With the generated AOG $\mathcal{OG}_0$ as the input, the three machine learning architectures estimate the POG $\hat{\mathcal{G}}_{t_\text{pred}}$, which are then compared with the ground truth POG $\mathcal{G}_{t_\text{pred}}$. The overview of the results can be seen in the TABLE III. and it prove that the machine learning algorithms can predict the future behaviour of the traffic participants.

TABLE III.   ERROR OF THE THREE MACHINE LEARNING ARCHITECTURES FOR THE VIDEO DATASET

| Architecture | $\bar{\epsilon}_{low}$ | $\bar{\epsilon}_{mid}$ | $\bar{\epsilon}_{high}$ |
|---|---|---|---|
| Architecture I | 0.0576 | 0.0348 | 0.0287 |
| Architecture II | 0.0802 | 0.0761 | 0.0524 |
| Architecture III | 0.1592 | 0.1479 | 0.0803 |

## V.   CONCLUSION

In this work, three different machine-learning architectures for an efficient estimation of the probabilistic space-time representation of the traffic scenario termed as the Predicted Occupancy Grid are presented. The machine learning algorithms used in the architectures are the Stacked Denoising Autoencoders, Random Forests and DeconvNet. Of the three machine learning architectures, a new architecture which uses a combination of two Stacked Denoising Autoencoders and reduced number of Random Forests to estimate the Predicted Occupancy Grid is also presented in this paper. A complex four way intersection with varying number of traffic participants is used for the validation. The simulations are carried out in the Matlab and TensorFlow environment. The results of the simulation using the 9000 test scenarios prove that the machine learning architectures are capable of predicting the evolution of traffic scenarios. The three architectures are compared both in terms of accuracy and their computational effort. The validation is also performed with the video dataset generated using the car-bicycle crash scenarios available in YouTube.

REFERENCES

[1]   H. Winner *et al*., "Handbook of Driver Assistance Systems," in *Springer,* 2016.
[2]   S. Lefèvre, C. Laugier and J. Ibañez-Guzmán, "Exploiting map information for driver intention estimation at road intersections," in *IEEE Intelligent Vehicles Symposium* (IV), Baden-Baden, pp. 583-588, 2011.
[3]   C. Tay, "Analysis of dynamic scenes: application to driver assistance", in PhD thesis, Instiut National Polytechnique de Gernoble, France, 2009.
[4]   D. Meyer-Delius, C. Plagemann, and W. Burgard, "Probabilistic situation recognition for vehicular traffic scenarios," in *IEEE International Conference on Robotics and Automation*, pp. 459–464, 2009.
[5]   P. Nadarajan and M. Botsch, "Probability Estimation for Predicted-Occupancy Grids in Vehicle Safety Applications Based on Machine Learning," in *IEEE Intelligent Vehicles Symposium, Gothenburg*, pp.1285-1292, 2016.
[6]   A. Armand *et al*., "Detection of Unusual Behaviours for Estimation of Context Awareness at Road Intersections," in *Workshop on Planning, Perception and Navigation for Intelligent Vehicles*, Tokyo, pp. 313-318, 2013.
[7]   P. Nadarajan, M. Botsch and S. Sardina, "Predicted-occupancy grids for vehicle safety applications based on autoencoders and the Random Forest algorithm," in *International Joint Conference on Neural Networks* (IJCNN), Anchorage, pp. 1244-1251, 2017.
[8]   L. Meng et al., "Research on denoising sparse autoencoder," in *International Journal of Machine Learning and Cybernetics*, pp. 1-11, 2016.
[9]   P. Vincent et al., "Extracting and Composing Robust Features with Denoising Autoencoders," in *Proceedings of the Twenty-fifth International Conference on Machine Learning* (ICML), pp. 1096-1103, 2008.
[10]   B. Schölkopf, J. Platt and T. Hoffmann, "Modeling human motion using binary latent variables," in *Advances in Neural Information Processing Systems*, MIT Press, pp. 1345-1352, 2007.
[11]   V. Badrinarayanan, A. Kendall, and R. Cipolla, "Segnet: A deep convolutional encoder-decoder architecture for image segmentation," in *arXiv preprint* arXiv*: 1511.00561*, 2015.

[12] D. Keysers et al., "Deformation Models for Image Recognition," in *IEEE Transactions on Pattern Analysis and Machine Intelligence*, vol. 29, no. 8, pp. 1422-1435, 2007.
[13] A. Chaulwar and M. Botsch, "Planning of safe trajectories in dynamic multi-object traffic-scenarios," in *Traffic and Logistics Engineering*, Vol. 4, No. 2, pp. 135-140, 2016.
[14] L. Breiman and E. Schapire, "Random Forests," in *Machine Learning*, Vol. 45, pp. 5-42, 2001.
[15] P. Geurts, D. Ernst and L. Wehenkel, "Extremely Randomized Trees," in *Machine Learning*, Vol. 63, pp. 3-42, 2006.
[16] H. Noh, S. Hong and B. Han, "Learning Deconvolutional Network for Semantic Segmentation," in *Proceedings of the IEEE International Conference on Computer Vision*, pp. 1520-1528, 2015.
[17] V. Dumoulin and F. Visin, "A Guide to Convolution Arithmetic for Deep Learning," in arXiv: 1603.07285, 2016.

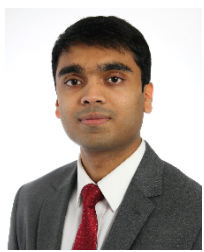

**Parthasarathy Nadarajan** is a Ph.D. candidate at the RMIT University, Melbourne, Australia and received the master degree in Automotive Engineering from Technische Hochschule Ingolstadt, Germany.

He is currently working as a research assistant at the vehicle safety research center CARISSMA at Technische Hochschule Ingolstadt, Germany since 2015. His research interests are in integrated vehicle safety and statistical learning.

Parthasarathy Nadarajan is a member of the International Neural Network Society.

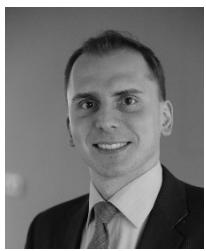

**Michael Botsch** received the diploma and doctoral degrees in electrical engineering, both with honors, from Technische Universität München, Munich, Germany, in 2005 and 2009.

He worked for five years in the automotive industry as Development Engineer at Audi AG in the field of active safety systems. In October 2013 he was appointed Professor for Vehicle Safety and Signal Processing at Technische Hochschule Ingolstadt in the Department of Electrical Engineering and Computer Science. He is the Associate Scientific Director of the vehicle safety research center CARISSMA at Technische Hochschule Ingolstadt. His research interests are in signal processing and automotive applications.

Prof. Botsch is a member of IEEE and VDE.

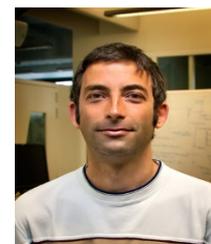

**Sebastian Sardina** obtained his Bachelor in Computer Science at the South National University, Argentina, and his PhD at the University of Toronto, Canada, before joining RMIT.

His research is mainly concerned with representation and reasoning in artificial intelligence for dynamic systems. His work spans several sub-fields of artificial intelligence, including automated planning, reasoning about action and change, agent-oriented programming, and reactive synthesis. He also has a strong interest in nonstandard forms of planning, including behavior composition synthesis of devices and agents.

Assistant Professor Sebastian Sardina is a regular member of the Senior Program Committee for IJCAI and AAMAS and of the Program Committee of almost every major AI conferences since 2006.